\documentclass{article} 
\usepackage{graphicx} 
\usepackage{microtype}
\usepackage{hyperref}
\usepackage{url}
\usepackage{booktabs}
\usepackage{multirow}

\usepackage{fullpage}
\usepackage{natbib}

\title{Reasoning about concepts with LLMs: Inconsistencies abound}

\author{Rosario Uceda-Sosa, Karthikeyan Natesan Ramamurthy, Maria Chang \& Moninder Singh \\
IBM Research\\
Yorktown Heights, NY 10598 USA \\
\texttt{\{rosariou,knatesa\}@us.ibm.com,maria.chang@ibm.com,moninder@us.ibm.com}
}
\date{}

\begin{document}

\maketitle

\begin{abstract}
The ability to summarize and organize knowledge into abstract concepts is key to learning and reasoning. Many industrial applications rely on the consistent and systematic use of concepts, especially when dealing with decision-critical knowledge. However, we demonstrate that, when methodically questioned, large language models (LLMs) often display and demonstrate significant inconsistencies in their knowledge.

Computationally, the basic aspects of the conceptualization of a given domain can be represented as Is-A hierarchies in a knowledge graph (KG) or ontology, together with a few properties or axioms that enable straightforward reasoning. We show that even simple ontologies can be used to reveal conceptual inconsistencies across several LLMs. We also propose strategies that domain experts can use to evaluate and improve the coverage of key domain concepts in LLMs of various sizes. In particular, we have been able to significantly enhance the performance of LLMs of various sizes with openly available weights using simple knowledge-graph (KG) based prompting strategies.  
\end{abstract}

\section{Introduction} \label{sec:introduction}

Conceptualization is a key cognitive ability that enables abstract thinking. Through concepts we communicate and learn complex knowledge by generalizing from instances and applying those learned principles to new situations. Conceptualization is at the base of symbolic reasoning and allows us to plan ahead and innovate beyond our physical experience.   

For example, children can easily conceptualize `chair' to the point of identifying new instances of chairs they haven't seen before and they apply the principle of `not putting your feet on a chair' to all chairs they may encounter in the future. Furthermore, when they learn `armchair', they understand it is a type of chair (Is-A hierarchy) and that whatever principles we apply to `chair' also apply to its sub-concept `armchair'. Not only children learn the concepts themselves, but they learn their associated Is-A hierarchies and how to reason about them {\em consistently}. 

Such consistent display, demonstration and reasoning using concepts is critical in several industrial applications where LLMs are used. Take, for example, a customer-facing chatbot in a property and casualty insurance company, it has to consistently demonstrate its knowledge of relations between various concepts in the ontology \citep{koutsomitropoulos2017standards}: if a `vehicle' is an `insurable object' that is covered according to a `policy', the LLM should consistently know that, say, `cruiser motorcycle', `van' or `scooter'  are vehicles but a `child's tricycle' is not considered a vehicle under the policy. Any inconsistency in behavior in identifying more specific sub-concepts (related by the IsA or subConceptOf relation) of `vehicle' could lead to a lack of trust in the system and downstream harm to the users.

\begin{figure}[t]
\begin{center}
\includegraphics[width=\textwidth]{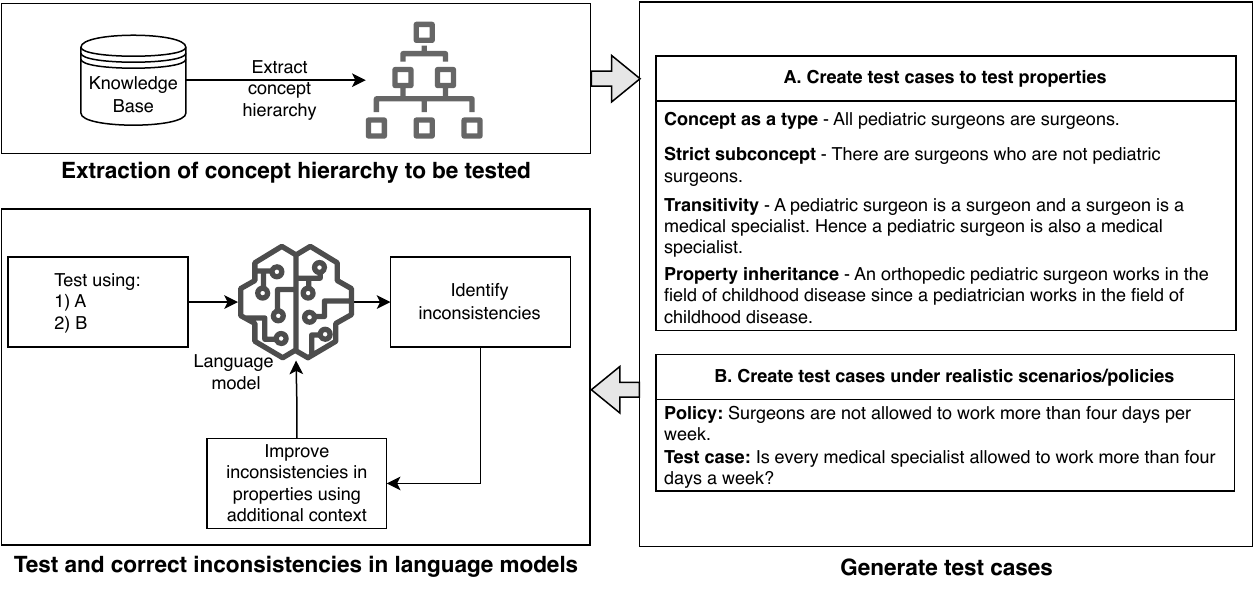}
\end{center}
\caption{Our proposed approach to test and correct for inconsistencies in an LLM's knowledge of concept hierarchies and in its application to realistic scenarios.}
\label{fig:outline}
\end{figure}

It is this consistent use of and reasoning about a concept hierarchy by LLMs that we propose to evaluate and discuss. That is, provided that an LLM has already some knowledge about concepts in a given domain and maybe some of their subConceptOf relations, we ask ourselves, is this knowledge consistently displayed in answering direct questions? Can we correct any inconsistencies with additional context? Can we leverage this knowledge {\em consistently} in simple reasoning tasks, for example, reason that a `cafe racer' and a `naked bike' are both types of motorcycles and that all properties of a motorcycle (like having a maximum capacity for two passengers) apply to both of them, as well as all other sub-concepts of motorcycle? How consistent is the LLM's  performance when answering these questions? We provide the outline of our proposed method to test and correct inconsistencies in Figure \ref{fig:outline}. We propose a three step process, starting with (1) The extraction of a concept hierarchy to be tested from a knowledge base, (2) the creation of various test cases to sieve inconsistencies via direct questioning (Figure \ref{fig:outline}A) and also reasoning about these concepts under realistic scenarios (Figure \ref{fig:outline}B). Finally, we (3) test the language model to identify inconsistencies and reduce them using additional context.

We leverage knowledge graphs (KGs) as a way to systematically define a concept hierarchy and the set of its entailments. The relation between KGs and LLMs is at the heart of a neuro-symbolic approach to AI. KGs provide structured, factual information in an algorithmic, traceable way, while LLMs offer advanced natural language understanding and generation. As interest on adapting LLMs to specialized domain vocabularies is growing \citep{SkilledLLM_1,SkilledLLM_2}, the integration of these complementary technologies holds the potential for creating more accurate and reliable AI systems, specially in applications requiring both precise information and sophisticated language capabilities \citep{alkhamissi2022review}. In this spirit, we will also ask here is about the KG-LLM integrations which enable key reasoning about concepts (as defined in Section~\ref{sec:defn_concept}. 

Our main contributions are: (1) We devise methods for using ontologies to assess the consistency and coverage of conceptualization in LLMs - this is done by creating test cases based on the knowledge graphs (KGs) or ontologies in an automated manner, (2) we demonstrate that several well-known LLMs with openly available weights demonstrate many inconsistencies in their knowledge, even with very rudimentary, small ontologies, and (3) we show that using simple prompting approaches we can reduce these inconsistencies and improve the coverage of domain concepts in several LLMs with openly available weights.

Our paper is structured as follows. We start with a working definition of conceptualization (Section \ref{sec:defn_concept}), then extract a sample ontology from Wikidata for our evaluation (Section \ref{sec:sample_ontology}). We define the inconsistencies we look for in LLMs (Section \ref{sec:logical_onto_inconsistencies}), and discuss a use case where we test the consistency in reasoning performed by LLMs for this ontology (Section \ref{sec:use_case}).  The results of our evaluation are discussed in (Section \ref{sec:eval}), followed by related work (Section \ref{sec:rel_work}) and conclusions/future directions (Section \ref{sec:future_dir}). Additional experimental details and results for one more domain (finance) are presented in the appendix. The datasets needed to reproduce our results along with prompts that we use are included in the supplemental material (uploaded separately).

\section{A working definition of conceptualization for KGs}
\label{sec:defn_concept}

We define a concept $C$ as a set of its instances. For example, `Medical Specialist' describes all the people whose professional occupation is a medical specialty. Subconcepts like `Surgeon' or `Pediatrician' represent subsets of medical specialists. A {\em subConceptOf} (also known as IsA) hierarchy of concepts is the simplest incarnation of an ontology, where every node represents a concept and the directed edges represent the subConceptOf relationship. This directed graph reflects a `mental picture' of the domain that users would expect to be stable and consistent.  

Here, we consider the key computational properties of conceptualization shown below: 

\begin{itemize} \label{concept_properties}
\item {\bf Concept as a type}. If $A$ is a subConceptOf $B$ then every instance of $A$ is an instance of $B$. Paraphrasing, every $A$ is also a $B$, or an $A$ is a type of $B$. E.g., all pediatric surgeons are surgeons. 
\item {\bf Strict subconcept property}. When the subConceptOf relation is strict, there are instances of $B$ that are not instances of $A$. E.g., there are surgeons who are not pediatric surgeons. 
\item {\bf Transitive property}. The relation subConceptOf is transitive. \textit{I.e.}, if $A$ is an instance of $B$, and $B$ is an instance of $D$, then $A$ is an instance of $D$. E.g., Given that a pediatric surgeon is a surgeon and a surgeon is a medical specialist, a reasonable user would infer that a pediatric surgeon is also a medical specialist. 
\item {\bf Subconcept property inheritance}. Every property that $B$'s have, $A$'s also have. E.g., if we assert that 'medical specialists must be board certified', we would also expect that surgeons and pediatric surgeons need to be board certified. 
\end{itemize}

There are other properties (axioms) that apply to conceptualizations (like reflexivity), but we consider that the four properties above sum up the behavior that most users would expect when reasoning about such a graph. For example, the subConceptOf property inheritance is very common in reasoning about rules and constraints, since it allows us to express them in the abstract, as in the example above. Also, when we apply the transitive property to a graph, we are effectively adding implicit edges to those in the graph that are explicit. This is usually called the {\em deductive closure} (with respect to a set of axioms) of the graph.

Even though this is an informal discussion about concepts and how most people would reason about them, we must remark that in the first property above we are equating the set theoretical definition of a concept (i.e., the set of its instances) with type theory (a concept is also a type). Most people won't have trouble understanding the context in which the term `surgeon' is used, and we expect that an LLM would do likewise.

\section{A Wikidata-based sample ontology}
\label{sec:sample_ontology}

\begin{figure}[h]
\begin{center}
\includegraphics[width=\textwidth]{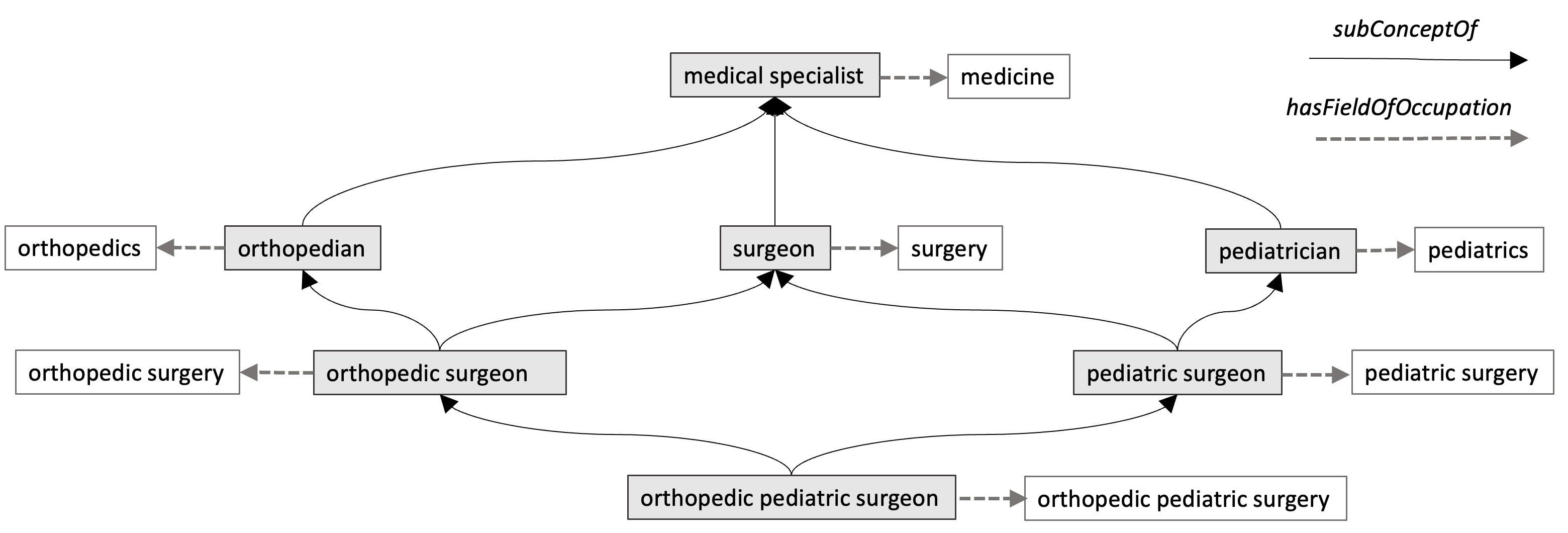}
\end{center}
\caption{A concept hierarchy snapshot: medical specialists and their specialities.}
\label{fig:pediatrician_hierarchy}
\end{figure}

To systematically evaluate the conceptual consistency in LLMs, we start with a small ontology fragment automatically extracted from Wikidata~\citep{Wikidata1,Wikidata2,Wikidata3}, the reason being that it is a well known KG, agreed upon by thousands of Wikidata contributors. We chose a common vocabulary, namely, medical specialties and specialists (surgeon, pediatrician, etc.) as shown in Figure~\ref{fig:pediatrician_hierarchy}. For the sake of conciseness, we have not drawn all of the subConceptOf relations between the medical specialties, but they are part of the underlying KG. 

We tried this small vocabulary with openly available LLMs, and we found that all of them answered correctly questions about the edges in over 90\% of our dataset question clusters, designed to test the edges and paths of the knowledge graph (see Section~\ref{sec:eval}). This fact alone demonstrates that the LLMs 'knew' of the graph vocabulary. Also, it is worth noting that our results are not domain specific, as we have obtained similar ones in other domains. In Appendix~\ref{sec:finance} we show an even simpler example from personal finances. 

To extract the ontology, we provided two seed concepts, the main entity, `medical specialist' (\href{https://www.wikidata.org/wiki/Q3332438}{Q3332438}), and one sample property to test the property inheritance, `medical specialty' (\href{https://www.wikidata.org/wiki/Q930752}{Q930752}). We get a graph with 3130 nodes (medical specialists and their occupations). 

From this initial graph, it is straightforward to extract the graph segment that includes their instances (\href{https://www.wikidata.org/wiki/Property:P31}{P31}) and subclasses (\href{https://www.wikidata.org/wiki/Property:P279}{P279}). Given that the differences between both of them are arbitrary in a higher order graph like Wikidata (where entities can be instances of a class and yet have instances themselves), we subsume them into the subConceptOf relation.

Reasoning about these concepts means being able to assert their deductive closure, i.e., the virtual, implied edges by the subConceptOf relation, say `orthopedic pediatric surgeon' and `surgeon'. This deductive closure is calculated by applying the axioms above to the existing graph and creating the dataset that we describe below.  

\section{Logical and ontological inconsistencies and question clusters}
\label{sec:logical_onto_inconsistencies}

The simplest form of logical contradiction is to both assert and deny the exact same fact, e.g., ``a cardiologist is a medical specialist'' and ``a cardiologist is not a medical specialist''. Other, more common, forms of logical inconsistency consist in asserting a fact and denying one of its (more or less immediate) consequences with respect to a given set of axioms or properties. For us, this set is described by the conceptualization properties in Section~\ref{sec:defn_concept} above. This also sits well with the formal definition of consistency discussed in~\cite{InconsistentKnowledge} -- ``in knowledge-based systems the notion consistency of knowledge is often understood as a situation in which a knowledge base does not contain contradictions.'' 

We test these properties through a set (or cluster) of simple queries with yes/no answers which are automatically generated from the KG. The questions in each cluster map to statements that must collectively be true or false (depending on how they are designed). A set of mixed answers reveals an inconsistency with respect to the conceptualization properties above. However, there is the possibility that an LLM answers `no' for an entire cluster, when the correct answer is `yes'. In that case, it could be that there's knowledge missing in the LLM. We call these {\em incomplete clusters} instead of {\em inconsistent clusters} (where answers are a mix of both `yes' and `no'). As we see in Section~\ref{sec:eval} the former are exceedingly rare (or non existent in several LLMs), meaning the LLMs tested `know' these concepts. 

While it is not possible to algorithmically sieve all the knowledge in LLMs, even using standard heuristics that exist to determine sets of unsatisfiable statements (statements which cannot possibly be all true at the same time and thus reveal an inconsistency), our approach allows an end user to define mission critical concept hierarchies and test them to ensure consistent responses for them. These graphs are small and test the key properties of concepts, as the general problem of identifying minimal unsatisfiable sets in KBs (equivalent to inconsistent clusters) is NP-hard ~\citep{InconsistentKBs,ImplicitIncoherentInconsistent,ContradictoryRelations}.

Before we describe these question clusters, we emphasize that all LLMs tested have responded correctly to some of their individual questions in over 98\% of the clusters (that is, we have very few incomplete clusters), even with a simple prompt. This means that the LLMs can process both the vocabulary as well as the linguistic forms shown here. We provide the full cluster datasets for the medical specialty domain, as well as the smaller financial domain in the supplementary materials.

\paragraph{Edge clusters} We test the first two properties in Section~\ref{concept_properties} with edge clusters. 

The first type of cluster is called a {\em positive edge cluster}. As the name indicates, it tests whether an edge exists or not in the KG, using various expressions in order to give the LLM linguistic flexibility and robustness. Take, for example, the "surgeon subconceptOf medical specialist". The corresponding edge cluster would be made of the following questions, for which we would expect all answers to be `yes' from the LLM.

\begin{itemize} \label{sample_edge_cluster}
    \item Is a surgeon a medical specialist? 
    \item Is a surgeon a type of medical specialist? 
    \item Is every surgeon a medical specialist? 
    \item Is a surgeon also a medical specialist? 
\end{itemize}

If an LLM answers all these questions in the negative, it is possible that it hasn't been trained or doesn't know this particular edge (i.e., it's an incomplete cluster). However, if say, all questions except the third question are answered `yes', there is obviously an inconsistency in the LLM knowledge. If every surgeon is NOT medical specialist, it cannot be that that a surgeon IS a medical specialist or that a surgeon is a type of medical specialist. That is, the answers to the questions imply an unsatisfiable set of statements. These questions are very simple and, in theory, it would be possible to increase the variations in questions in each of the clusters, but it would only likely make the model responses more inconsistent according to our definition, and so the performance we report here is an optimistic estimate.

We also formulate edge clusters whose answer should be negative. In particular, {\em inverse edge clusters} are used to test the {\em strict subconcept property} above, when a concept A is strictly contained in its parent B, meaning that there are instances of B that are not instances of A. For example, for the inverse of the cluster above, a subset of the questions we ask would be:
\begin{itemize}
\item Is every medical specialist a surgeon?
\item Is a medical specialist a type of surgeon?
\end{itemize} 

These can be generated automatically by comparing the instances (\href{https://www.wikidata.org/wiki/Property:P31}{P31}) of both A and B and checking there is no 'same as' property (\href{https://www.wikidata.org/wiki/Property:P460}{P460}) between them (which is exceedingly rare).

The third type of edge cluster, the {\em negative edge cluster} also tests the first set theoretic property of conceptualization, but in the negative. We automatically select other nodes in the hierarchy that do not have a subConceptOf relation by choosing pairs far away in the hierarchy, for example, [cardiologist, dermatologist] or [surgeon, hypnotherapist]. The questions are linguistically formulated as in the {\em positive edge clusters}. In this case, a subset of the questions we have would be:
\begin{itemize}
\item Is a surgeon a hypnotherapist?
\item Is a surgeon a type of hypnotherapist?
\end{itemize}

It is important to notice that some of the questions where the LLM disagrees with the ground truth answer in our dataset may be technically correct. An LLM may object to one of these particular linguistic forms and may make a well reasoned argument for its answer. For example, when asking ``is an orthopedic pediatric surgeon an infection control physician?'' the language model (mixtral-8x7b-instruct in this case), instead of a `yes' or `no' answer, offers an explanation for a non-committal answer: ``it is possible that an orthopedic pediatric surgeon may work in the field of infection control, however this is not their primary field of occupation, which is orthopedic surgery and pediatric surgery''. This answer is technically correct, but not {\em consistent} with the answers to the majority of similar questions such as, ``is a orthopedic pediatric surgeon a infectious disease physician ?'' or ``is a orthopedic pediatric surgeon a hepatologist?'', and dozens of others this same LLM answers simply `no'. Given that we are not testing the LLM knowledge, but its consistency, we still have to mark this answer -when compared to the majority of similar answers- inconsistent. 

\begin{figure}[h]
\begin{center}
\includegraphics[width=\textwidth]{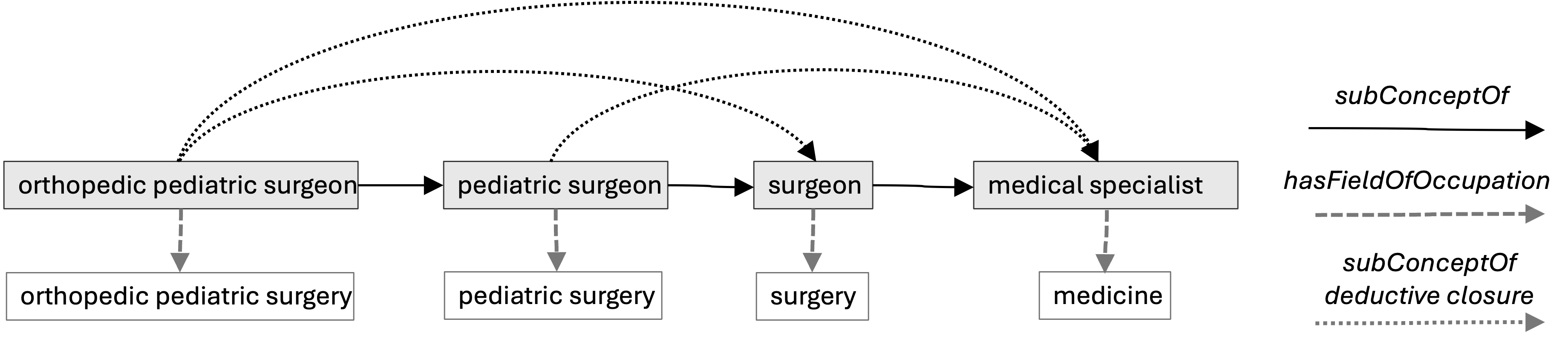}
\end{center}
\caption{Deductive closure between orthopedic pediatric surgeon and medical specialist.  }
\label{fig:hierarchy paths}
\end{figure}

\paragraph{Path clusters} This second type of cluster tests the transitivity of the subConceptOf relation in properties~\ref{concept_properties} by querying a sequence of edges in a given path. Using the same linguistic forms as before, we ask about the deductive closure of a path (the curved arrows in fig~\ref{fig:hierarchy paths}). In our sample graph there are 4 such paths. Two of these are, {\em [orthopedics pediatric surgeon, pediatric surgeon, surgeon, medical specialist]}, as shown, and {\em [orthopedics pediatric surgeon, orthopedics surgeon, orthopedian, medical specialist]}. 

\paragraph{Property hierarchy clusters} The last type of cluster tests the fourth property of conceptualization, {\em subconcept property inheritance}. This is an core feature of conceptualization that affords abstract reasoning. For example, consider the questions below: 

\begin{itemize}
\item is the field of occupation of a surgeon surgery?
\item is an orthopedic pediatric surgeon a surgeon ?
\item is the field of occupation of a orthopedic pediatric surgeon surgery?
\end{itemize}

If the field of occupation of a surgeon is surgery, and an orthopedic pediatric surgeon is a surgeon, we would expect that the field of occupation of an orthopedic pediatric surgeon is also surgery. Of course, a more specific answer is that orthopedic pediatric surgery is the occupation of an orthopedic pediatric surgeon, but the fact remains that all of the models tested answer the above cluster in the affirmative in the majority of cases. Again, it is the matter of consistency that concerns this study. 

\section{Demonstrative use case}
\label{sec:use_case}

Why is it important to ensure that an LLM can consistently answer seemingly simple questions about the edges of a given KG? Imagine a set of policies, rules or processes that a health care network or an insurance company wants to define and use in an AI application. Take, for example:
\begin{enumerate}
    \item ``Only pediatric surgeons can perform surgery on patients younger than 18 years old.''
    \item ``Only surgeons are required to work no more than four days per week.''
\end{enumerate}

Not only the policy designers would expect to define and manage these rules using abstract concepts, but the users of the application would expect to query these policies using more specific vocabulary related to their case. Somehow, the application should be able to \textit{understand} whether a pediatric surgeon or a pediatrician satisfy either policy.   

We have created a small dataset of 10 scenarios with simply worded policies that apply to the medical specialists in our sample knowledge graph (included in our supplemental materials). Each scenario is tested with two types of questions. The first one is ``Does the policy apply to every \{specialist\}?'' where \{specialist\} is substituted by one of the 7 terms in our sample graph (`pediatrician', `surgeon', `orthopedic surgeon' and so on). The second type of question mimics the policies above, using the same type of term substitution. The queries corresponding to the policies above are:  

\begin{enumerate}
\item Is every \{specialist\} allowed to treat or operate on patients younger than 18 years old?
\item Is every \{specialist\} allowed to work more than four days per week?
\end{enumerate} 

Knowledge of our sample graph, or the equivalent implicit knowledge, is required to answer these straightforward questions correctly, which shows that concept hierarchies lie at the base of this type of industrial applications. However, as we see in Table~\ref{tab:realistic_scenarios} many of the LLMs with openly available weights get many individual answers wrong, even though they also get some answers right. It is worth noting that there are no `incomplete' scenarios (where every individual question is incorrect) here. So, we ask ourselves again, what happened? Is it lack of specific knowledge (one edge or one node) or lack of overall consistency in the knowledge? Why do the LLMs fail to answer correctly in some cases and not in others? Can we pinpoint the specific \textit{holes} in the knowledge so it can be corrected?  

To dig deeper into these questions, we need to generate a dataset to test systematically the knowledge graph directly, as we have discussed in Section~\ref{sec:logical_onto_inconsistencies}.   

\begin{table}[htbp]
  \centering
  \caption{Evaluation of 10 policy-based scenarios (14 questions per scenario).}
  \setlength{\tabcolsep}{2pt}
    \begin{tabular}{lcc}
    \toprule
    \multicolumn{1}{l}{\multirow{3}[1]{*}{LLM name}} & \multicolumn{1}{c}{\% incorrect} & \multicolumn{1}{c}{\% inconsistent} \\
    & \multicolumn{1}{c}{individual} & \multicolumn{1}{c}{scenarios (10)} \\
    & \multicolumn{1}{c}{answers (140)} & \multicolumn{1}{c}{} \\
    \midrule
    google/flan-t5-xl \citep{flan-t5} & 65.71 & 100 \\
    google/flan-t5-xxl \citep{flan-t5} & 24.28 & 90 \\
    google/flan-ul2 \citep{flan-ul2} & 15    & 70 \\
    meta-llama/llama-2-13b-chat \citep{llama} & 22.8  & 80 \\
    meta-llama/llama-2-70b-chat \citep{llama} & 15    & 60 \\
    tiiuae/falcon-180b \citep{falcon} & 15    & 60 \\
    mistralai/mistral-7b-instruct-v0-2 \citep{mistral} & 13.57 & 60 \\
    mistralai/mixtral-8x7b-instruct-v0-1 \citep{mixtral} & 13.57 & 40 \\
    thebloke/mixtral-8x7b-v0-1-gptq \citep{thebloke-mixtral} & 35    & 100 \\
    \bottomrule
    \end{tabular}%
  \label{tab:realistic_scenarios}%
\end{table}%

\section{Evaluation and coverage improvement} 
\label{sec:eval}

The three types of clusters described above are designed to highlight the inconsistencies of the LLM knowledge. We automatically extract them from the topology of the test KG above, producing 119 clusters, with 96 edge clusters (the high number is due to the fact that we have negative and inverse edge clusters representing edges NOT in the graph). More details are provided in Appendix~\ref{sec:appendix A}, and additional results for a different domain ontology are presented in Appendix~\ref{sec:finance}.

We test this graph in 9 openly available models 
(see Table \ref{tab:realistic_scenarios} for model information)
using a simple prompt with 11 sample questions from the medical domain. These models are hosted in our own organization's infrastructure. The prompt used is provided in the supplementary materials. We ask for yes/no answers which can be automatically tallied. A `yes' answer means that, for every possible instance, the question can always be answered in the affirmative. Otherwise, the answer should be `no', as it doesn't hold for the concept (i.e., all its instances). The results are displayed in Table~\ref{tab:eval_prompt_naive}. For conciseness' sake, we have added all the edge clusters together. A few facts worth noting. First, we notice in the leftmost column that there are very few incomplete edges (where all the individual responses in a cluster are wrong). This means that out of the 96 edge clusters, the vast majority of them are \textit{known} to the LLMs. Some LLMs have at least one correct answer in every single one of the clusters - no incomplete edges.  Second, we notice that notion of property inheritance is the most challenging, where all of the models fail over 36\% of the time. 

\begin{table}[htbp]
\small
  \centering
  \caption{Evaluation results by model using a simple prompting strategy.}
  \setlength{\tabcolsep}{2pt}
    \begin{tabular}{lccccc}
    \toprule
    \multicolumn{1}{l}{\multirow{3}[1]{*}{LLM name}} & \multicolumn{1}{c}{\% incomp.} & \multicolumn{1}{c}{\% incons.} & \multicolumn{1}{c}{\% incons.} & \multicolumn{1}{c}{\% incons.} & \multicolumn{1}{c}{\% all} \\
          & \multicolumn{1}{c}{edges} & \multicolumn{1}{c}{edges} & \multicolumn{1}{c}{paths} & \multicolumn{1}{c}{property} & \multicolumn{1}{c}{incons.} \\
          & \multicolumn{1}{c}{(96)}      &  (96)     &  (12)     & \multicolumn{1}{c}{inherit. (11)} & \multicolumn{1}{c}{(119)} \\
    \midrule
    google/flan-t5-xl & 4.17  & 40.62 & 16.66 & 36.36 & 41.18 \\
    google/flan-t5-xxl & 1.04  & 35.41 & 16.66 & 36.36 & 34.45 \\
    google/flan-ul2 & 4.17  & 26.04 & 33.33 & 54.54 & 32.77 \\
    meta-llama/llama-2-13b-chat & 0     & 13.54 & 16.66 & 36.36 & 15.97 \\
    meta-llama/llama-2-70b-chat & 3.13  & 22.91 & 16.66 & 45.45 & 26.89 \\
    tiiuae/falcon-180b & 0     & 17.7  & 16.66 & 36.36 & 19.33 \\
    mistralai/mistral-7b-instruct-v0-2 & 0     & 4.16  & 25    & 36.36 & 9.24 \\
    mistralai/mixtral-8x7b-instruct-v0-1 & 2.08  & 22.91 & 16.6  & 36.36 & 25.21 \\
    thebloke/mixtral-8x7b-v0-1-gptq & 1.04  & 32.29 & 16.66 & 36.36 & 31.93 \\
    \bottomrule
    \end{tabular}%
  \label{tab:eval_prompt_naive}%
\end{table}%

\begin{table}[htbp]
\small
  \centering
  \caption{Evaluation results by model with prompt augmented by context to improve consistency.}
  \setlength{\tabcolsep}{2pt}
  \setlength{\tabcolsep}{2pt}
    \begin{tabular}{lcccccc}
    \toprule
    \multicolumn{1}{l}{\multirow{3}[1]{*}{LLM name}} & \multicolumn{1}{c}{\% incomp.} & \multicolumn{1}{c}{\% incons.} & \multicolumn{1}{c}{\% incons.} & \multicolumn{1}{c}{\% incons.} & \multicolumn{1}{c}{\% all} & \multicolumn{1}{c}{\% improve.}\\
          
    & \multicolumn{1}{c}{edges} & \multicolumn{1}{c}{edges} & \multicolumn{1}{c}{paths} & \multicolumn{1}{c}{property} & \multicolumn{1}{c}{incons.}  & \multicolumn{1}{c}{(all}\\
          & \multicolumn{1}{c}{(96)}      &  (96)     &  (12)     & \multicolumn{1}{c}{inherit. (11)} & \multicolumn{1}{c}{(119)} & \multicolumn{1}{c}{incons.)}\\
    \midrule
    google/flan-t5-xl & 1.04  & 10.41 & 25    & 27.27 & 14.29 & \textbf{26.89} \\
    google/flan-t5-xxl & 1.04  & 10.41 & 0     & 0     & 9.24  & \textbf{25.21} \\
    google/flan-ul2 & 1.04  & 12.5  & 0     & 27.27 & 13.45 & \textbf{19.33} \\
    meta-llama/llama-2-13b-chat & 0     & 7.29  & 0     & 9.09  & 6.72  & \textbf{9.24} \\
    meta-llama/llama-2-70b-chat & 2.08  & 10.41 & 0     & 9.09  & 10.92 & \textbf{15.97} \\
    tiiuae/falcon-180b & 1.04  & 13.54 & 0     & 0     & 11.76 & \textbf{7.56} \\
    mistralai/mistral-7b-instruct-v0-2 & 0     & 6.25  & 0     & 0     & 5.04  & \textbf{4.20} \\
    mistralai/mixtral-8x7b-instruct-v0-1 & 0     & 9.37  & 0     & 27.27 & 10.08 & \textbf{15.13} \\
    thebloke/mixtral-8x7b-v0-1-gptq & 0     & 9.375 & 0     & 27.27 & 10.08 & \textbf{21.85}\\  
    \bottomrule
    \end{tabular}%
  \label{tab:eval_prompt_context}%
\end{table}%

Next, we look to enhance the performance of the initial prompt by adding to the queries a context with the propositionalization of the knowledge that was missed by all the models, i.e., we use the same context for all the models. This context is computed automatically, as our underlying dataset (included in supplemental file) maps the cluster questions into their corresponding assertions. For example, `is every orthopedic surgeon a surgeon?' is mapped to `every orthopedic surgeon is a surgeon'. This allows us to generate the context for queries on a second test. This `wholesale' approach to context augmentation yields roughly the same improvements as if we tailored the context to each individual model. 

With this simple prompt augmentation strategy, we obtain a sizable performance enhancement as shown in Table~\ref{tab:eval_prompt_context}. The rightmost column reflects the performance enhancement in the clusters, showing that now points of inconsistency have been reduced up to one third. It is worth noting that even this explicit knowledge doesn't eliminate inconsistency altogether.  

\section{Related Work}
\label{sec:rel_work}

Seminal work by \cite{petroni2019language} demonstrated that a language model could learn relational knowledge (i.e. facts one would expect to be found in a knowledge base) during pre-training. This raised the possibility that language models could serve as approximations for knowledge bases right out of the box. However, \cite{elazar2021measuring} used paraphrased querying to show that such knowledge could not elicited consistently/reliably. This led to the development of frameworks for measuring inconsistency in language models \citep{jang2021accurate, laban2023llms, sahu2022unpacking} as well as novel training setups with consistency-based loss \citep{elazar2021measuring}. The consistency issues found in LLMs have been identified as one of the key areas of future work needed to enhance LLMs so they share the same strengths -and consistency- as KBs \citep{alkhamissi2022review}.

Large language models have recently been shown to exhibit abilities akin to `reasoning' when prompted in certain ways. For example, chain-of-thought prompting ( \cite{wei2022emergent}) gets models to provide explicit steps it took to arrive at an answer. Nevertheless, it is not clear whether it actually demonstrates that the LLMs are actually reasoning \cite{ wei2022emergent, kojima2023large}. \cite{wang2023selfconsistency} explores the consistency of LLM results via chain-of-thought and studies ways of making such results more consistent. A nice survey on the current state of knowledge in reasoning in LLMs is provided by \cite{ huang2023reasoning}. Other work has looked what LLMs actually know \cite {yin2023large, srivastava2023imitation, sun2023headtotail} and have shown that LLMs exhibit are very weak in this regard, with performance sometimes barely surpassing random guessing \cite{srivastava2023imitation}.

Improving consistency and factual correctness of language models is related to ongoing work that aims to integrate external knowledge into LLMs, either from unstructured sources like retrieved documents or from structured knowledge bases \citep{feng2023trends,yang2024facts}. Approaches may be applied at different stages of the model lifecycle \citep{Pan_2024}: KGs may be used in pre-training \citep{yasunaga2022deep}, tuning \citep{zhang2024knowledge, cheng2023editing} or information from KGs can be incorporated directly into the prompt \citep{Andrus_Nasiri_Cui_Cullen_Fulda_2022, fatemi2024talk}. 

Our proposed approach differs from the above related works in that we perform analysis of consistency of knowledge of LLMs with respect to a small and targeted KG by automatically generating test cases. Our clusters can act as building blocks of satisfiability -or unsatisfiability-, so we can identify small portions of knowledge to edit or evaluate. Also, we do not require an externally annotated dataset, such as a QA benchmark. We also perform targeted editing of the LLM's knowledge using prompting. This is because in industrial applications, the domain expert requires consistency in a relatively small fragment of a specialized KG. For example, in a general KG, a bicycle is objectively a type of vehicle, but in our introductory insurance example, bicycles are typically not covered by vehicle insurance and so they cannot be considered vehicles per the insurance contract. This means that domain experts may need to edit the knowledge. While KG reasoning and editing in general may be useful like in  GraphRAG~\footnote{https://www.microsoft.com/en-us/research/blog/graphrag-unlocking-llm-discovery-on-narrative-private-data/} or \citep{reasoningOnGraphs}, we explore more targeted editing that can be systematically tested and verified to gain the trust of the domain experts and other relevant stakeholders. 

\section{Conclusions and future work}
\label{sec:future_dir}

Consistent conceptualization, especially when addressing mission critical data, is key in industrial applications. We have shown that inconsistencies creep in LLMs even when using common vocabulary and even after prompting the system with targeted content. There are some natural future directions that emerge from these insights.

The first looks to identify knowledge issues and systematically evaluate an LLM for them. This may be done by mapping the knowledge from a KG to richer, linguistically more challenging queries that users may realistically pose to the LLM. Using train-of-thought factoring of the user query into simpler queries, like the ones we produce, may help in this mapping.

The second could be to allow for questions that require non-committal answers and thereby handle ambiguous contexts. For example, the question `does a pediatric surgeon always work with children?' may have a `maybe' answer, as pediatric surgeons also work with teenagers. Part of establishing trust in the LLM is to ensure that ambiguous queries are properly, and consistently, dealt with.

\section*{Acknowledgements}
\label{sec:ack}
The authors thank Kush Varshney for his advice and support.

\bibliography{colm2024_conference}
\bibliographystyle{colm2024_conference}

\appendix
\section{Cluster dataset construction} \label{sec:appendix A}
Starting with the seed concept `orthopedic pediatric surgeon', we automatically generate a data set that comprises 109 clusters, with a total of 584 questions, which include 4 different linguistic forms per query, so we have approximately 146 \textit{semantically different} queries (some of the property clusters have 2 questions only per medical specialty). The size of the dataset is as follows: 

\begin{itemize}
\item 15 positive edge clusters.
\item 66 negative edge clusters. The number of these can be adjusted with a parameter. Obviously, in a small hierarchy, looking for unrelated nodes becomes harder if the top number is higher. 
\item 15 inverse edge clusters. 
\item 12 path clusters. 
\item 11 property inheritance clusters. 
\end{itemize}

Each cluster, regardless of type is made up of the following:

\begin{itemize}
\item Expected answer. `yes' or `no'.
\item Source. This is the source node in the directed graph
\item Target. The target node of an edge or a path cluster. 
\item Questions. These are generated from fixed linguistic patterns for subConceptOf and for property edges. For example: ``is a orthopedic pediatric surgeon a medical specialist ?",
``is a orthopedic pediatric surgeon a type of medical specialist ?", ``is every orthopedic pediatric surgeon a medical specialist ?" and ``is a orthopedic pediatric surgeon also a medical specialist ?" for an edge cluster with source `orthopedic pediatric surgeon'. 
\item Statements. The corresponding statements to the questions above:  ``a orthopedic pediatric surgeon is a medical specialist", ``a orthopedic pediatric surgeon is a type of medical specialist", ``every orthopedic pediatric surgeon is a medical specialist" and ``a orthopedic pediatric surgeon is also a medical specialist". These statements are used to create augmented context to improve the consistency of the LLMs.
\end{itemize}

Our full json dataset is provided in the supplementary file.

\section{The finance domain } \label{sec:finance}

To prove how pervasive the inconsistencies in LLMs are, we have tried a variety of domains, government agencies, corporate occupations and finance. In this case, we have created an overly simple graph of just one path without any property edges, and straightforward vocabulary, so we can test the first three properties of conceptualization. 

\begin{figure}[h]
\begin{center}
\includegraphics[width=\textwidth]{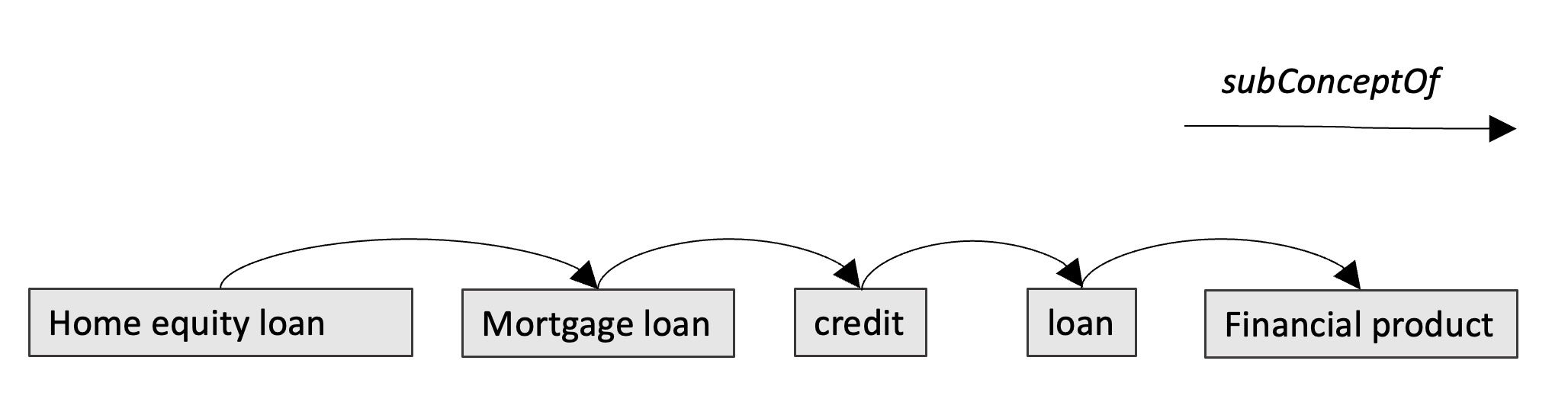}
\end{center}
\caption{Finance domain: home equity loan path }
\label{fig:financial_path}
\end{figure}

This dataset has a total of 75 edges and only 4 hierarchy clusters. No property inheritance clusters, as mentioned before. The results below may not be statistically significant but we include them here because they are revealing in a couple of ways. 

Even in this Hello World example, we get inconsistencies in similar percentages as above. Also, it is interesting that even after specifically adding the context in the prompt, we don't necessarily improve the performance in all models with respect to this simple path. 

Finally, the fact that after prompting some models degrade slightly in performance (probably without statistically significance), indicates that only prompting may not be the only answer. 

\begin{figure}[h]
\begin{center}
\includegraphics[width=\textwidth]{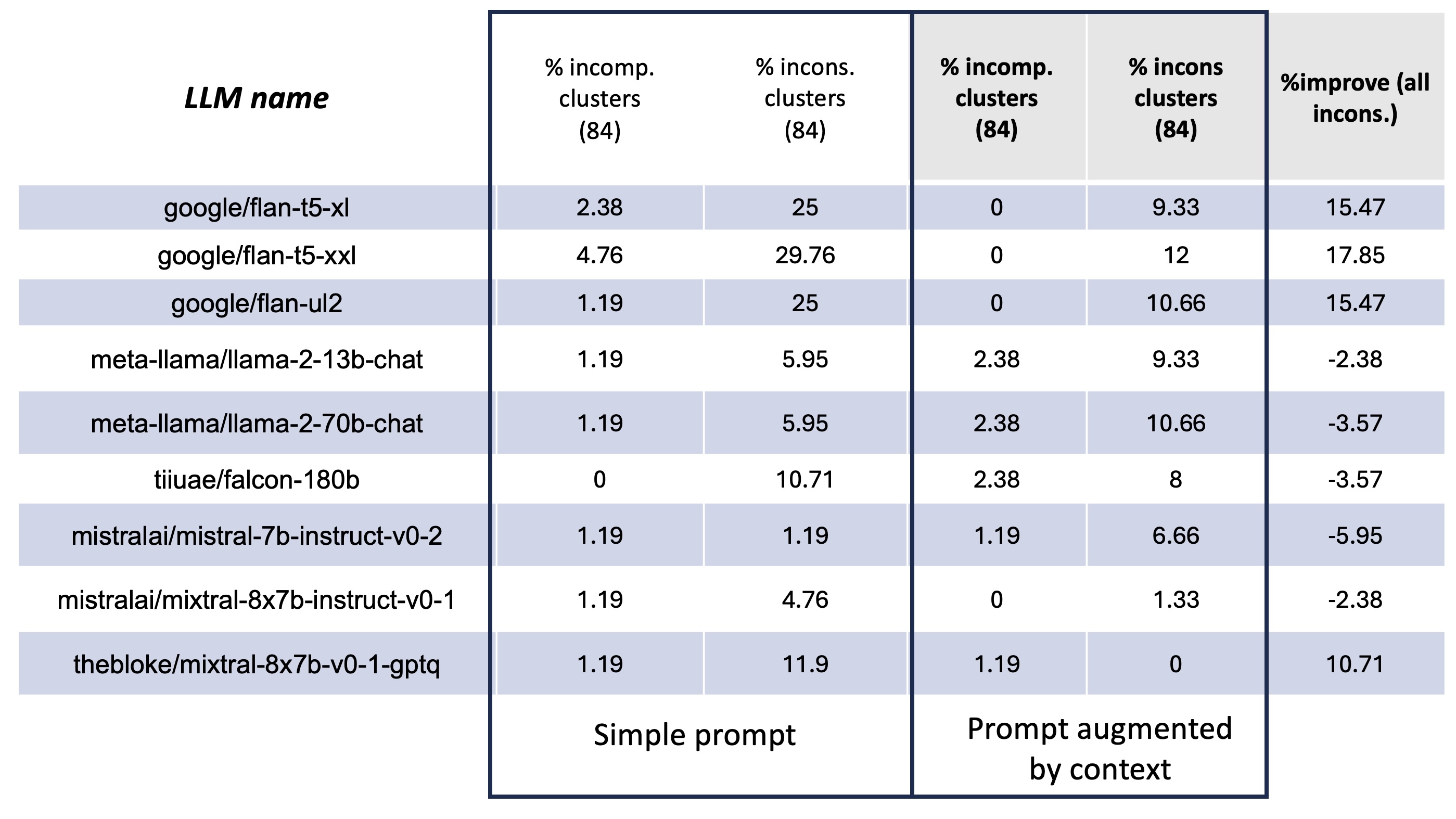}
\end{center}
\caption{Finance domain: eval with simple prompt and with context }
\label{fig:financial_path}
\end{figure}

\section{Ethics statement}
\label{sec:ethics_statement}
Our datasets were created by ourselves using publicly available wikidata ontologies. The content of our knowledge graphs is common knowledge and we do not involve any human subjects for data generation or validation.

One of the key motivations of our proposed approach is to enable users calibrate trust in LLMs and improve the consistency of LLMs in specific domains to make them more trustworthy. We believe that exhaustive testing using methods such as ours is necessary in any high-stakes application. A potential issue in using exhaustive testing methods such as what we propose is that a lot of inference calls need to be made to LLM and this increases their power consumption. However, this is mitigated by the fact that this needs to be done only for the domains of application in which the LLM is used. This testing will also reduce downstream harms for users that may happen due to inconsistent knowledge in the models.

\section{Reproducibility statement}
\label{sec:reproducibililty_statement}
We provide the dataset that we generated in the supplementary material. This can be used to test any model in the domains that we presented in the paper. We also discuss the methodology by which we created the dataset in sufficient detail in the paper. Any knowledgeable reader can use a similar methodology to test their own model in a domain of interest.

\end{document}